\documentclass[runningheads]{llncs}

\usepackage{eccv}

\usepackage{eccvabbrv}

\usepackage{graphicx}
\usepackage{booktabs}
\usepackage{colortbl}
\usepackage{multirow}
\usepackage{wrapfig}

\usepackage[accsupp]{axessibility}  

\usepackage[pagebackref,breaklinks,colorlinks,citecolor=eccvblue]{hyperref}

\usepackage{orcidlink}

\newcommand{\model}{World Volume-aware Multi-camera Driving Scene Generator}
\newcommand{\abb}{WoVoGen}
\newcommand{\W}{\mathcal{W}}

\begin{document}

\title{\abb{}: World Volume-aware Diffusion for Controllable Multi-camera Driving Scene Generation}

\titlerunning{WoVoGen}

\author{Jiachen Lu\inst{1} \and
Ze Huang\inst{1} \and
Zeyu Yang\inst{1} \and
Jiahui Zhang\inst{1} \and
Li Zhang\inst{1}\thanks{Li Zhang (lizhangfd@fudan.edu.cn) is the corresponding author.
}}

\authorrunning{J. Lu, et al.}

\institute{School of Data Science, Fudan University\\
\vspace{4mm}
\url{https://github.com/fudan-zvg/WoVoGen}}

\maketitle

\begin{figure*}[th]  
\centering  
\includegraphics[width=\linewidth]{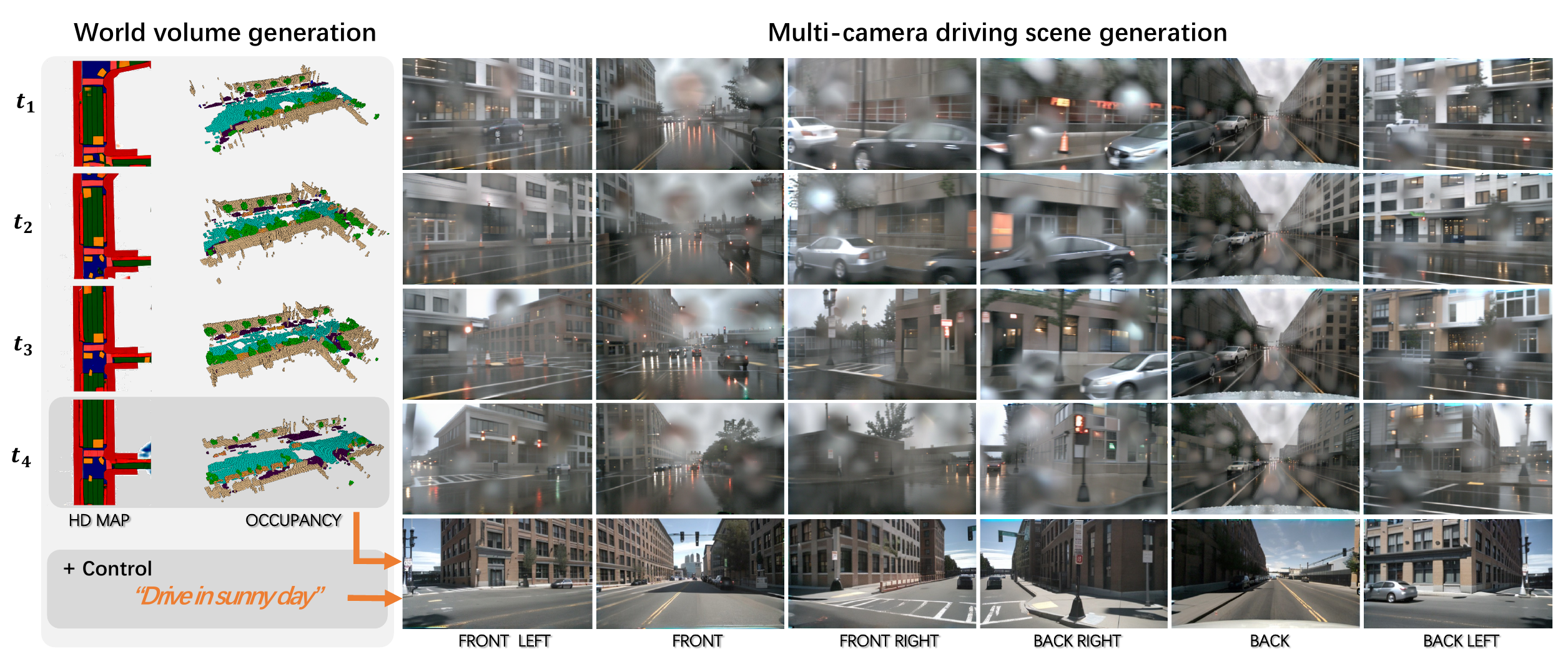} 
\caption{Our \abb{} is crafted to generate future world volumes (\ie, HD maps and occupancy) and high-quality multi-camera street-view images, with the input of past world-volumes.
The bottom row shows the weather-based control of \abb{}.
Specifically, with the predicted world volume at time $t_4$ and a weather description, the multi-camera images transit from rainy to sunny conditions, while maintaining the street layout.
}
\label{fig:video_teaser} 
\end{figure*}

\begin{abstract}
Generating multi-camera street-view videos is critical for augmenting autonomous driving datasets, addressing the urgent demand for extensive and varied data.
Due to the limitations in diversity and challenges in handling lighting conditions, traditional rendering-based methods are increasingly being supplanted by diffusion-based methods.
However, a significant challenge in diffusion-based methods is ensuring that the generated sensor data preserve both intra-world consistency and inter-sensor coherence.
To address these challenges, we combine an additional explicit world volume and propose \model{} (\abb{}). 
This system is specifically designed to leverage 4D world volume as a foundational element for video generation.
Our model operates in two distinct phases: 
(i) envisioning the future 4D temporal world volume based on vehicle control sequences, and 
(ii) generating multi-camera videos, informed by this envisioned 4D temporal world volume and sensor interconnectivity.
The incorporation of the 4D world volume empowers \abb{} not only to generate high-quality street-view videos in response to vehicle control inputs but also to facilitate scene editing tasks.
\end{abstract} 
\section{Introduction}
\label{sec:intro}
The burgeoning field of vision-based autonomous driving perception~\cite{li2022bevformer, radford2021learning, philion2020lift} underscores the need for high-quality multi-camera street-view datasets~\cite{caesar2020nuscenes}. With the notorious costs of labeling in autonomous driving datasets, there is a significant demand for generating high-quality multi-camera street-view videos that accurately mirror real-world 3D data distributions and maintain consistency across multiple sensors.

Recent {driving scene content synthesis} techniques can be categorized into two main groups: rendering-based~\cite{wu2023mars, yang2023unisim, guo2023streetsurf, xie2023s, chen2023periodic} and diffusion-based~\cite{swerdlow2023street, yang2023bevcontrol, gao2023magicdrive, wang2023drivedreamer, li2023drivingdiffusion} methods.
Rendering-based methods benefit from an explicit 3D or 4D world structure, ensuring stringent 3D consistency. Yet, this approach tends to offer limited diversity and requires considerable effort to produce multi-camera videos that comply with real-world lighting, weather conditions, \textit{etc}.
On the other hand, methods based on finetuned stable diffusion models boast high diversity and can easily generate images that follow real-world distributions. 
{Despite the noted advantages, diffusion-based approaches \cite{swerdlow2023street, yang2023bevcontrol, gao2023magicdrive, li2023drivingdiffusion, wang2023drivedreamer} that employ bounding boxes or HD maps as control signals offer a sparse representation of the driving scene. 
Consequently, the generated videos often lack critical information about the scene, leading to an incomplete understanding. This limitation underscores the necessity for a diffusion-based model that incorporates a dense representation of the world as a condition, ensuring a more comprehensive and accurate depiction of the driving environment.}

To overcome the limitations of diffusion-based methods, we introduce \model{} (\abb{}), a framework designed to endow diffusion-based generative models with an explicit 4D world volume. 
Our approach operates in two stages: initially, we envision a 4D world volume using a reference scene combined with a future vehicle control sequence. 
Subsequently, this volume guides the generation of multi-camera footage.
Concretely, our 4D world volume manifests as a dense voxel volume spanning four dimensions: time, height, length, and width, corresponding to the scope of the bird's-eye view (BEV) domain. 
This representation encapsulates the scene's comprehensive data, comprising object occupancy, high-definition maps, background details, and road attributes. 
Each voxel within this 4D construct is annotated with a class label, providing a rich, multi-faceted understanding of the environment.

In the preliminary stage, our method is to train an autoencoder model~\cite{van2017neural} that encodes a single-frame 3D world volume into a 2D latent representation. 
Subsequently, we stack these 2D latents along the temporal axis to form a 2D temporal latent series.
We further refine the UNet by introducing temporal versions of its residual and cross-attention blocks, which can effectively process the time-varying information, guided by the vehicle control sequence as the conditional context. Upon generating the future 2D temporal latent, we proceed to decode it back into the 4D world voxel volume using the autoencoder's decoder.
Having generated the future 4D world voxel volume, we employ a combination of CLIP and 3D sparse CNN to convert it into a 4D world feature. 
This feature is then transformed geometrically to sample 3D image volumes for each camera, corresponding to each time step. 
Subsequently, these 3D image volumes are condensed into 2D image features, which serve as conditional inputs for ControlNet~\cite{zhang2023adding}.
Alongside image features, we employ textual prompts as scene guidance akin to those used in Stable Diffusion~\cite{rombach2022high} to govern the overall scene conditions such as weather, lighting, location, and the scene at large. 
For more precise object location control within the scene, textual prompts are utilized with greater specificity. 
We map the 4D world volume labels onto each 2D pixel, utilizing the label names as textual conditions to objectively guide over each pixel's characteristics.
For achieving inter-sensor consistency, we concatenate surround-view images into a meta-image and leverage the diffusion model to learn the distribution of real-world multi-view image sequences. To ensure temporal coherence, we utilize the same temporal Transformer blocks previously described.

In summary, we make the following \textbf{contributions}:
\textbf{(i)} We propose \abb{}, a framework that leverages an explicit world volume to guide the diffusion model generation, ensuring intra-world and inter-sensor consistency.
\textbf{(ii)} We introduce a two-phase strategy: initially envisioning the future 4D world volume, followed by generating multi-camera videos based on this envisioned world volume.
\textbf{(iii)} Our system not only excels in producing street-view videos that exhibit consistency within the world and across various cameras, driven by vehicle control inputs, but also facilitates scene editing tasks.

\section{Related work}
\label{sec:related}

\begin{figure*}[th]  
\centering  
\includegraphics[width=\linewidth]{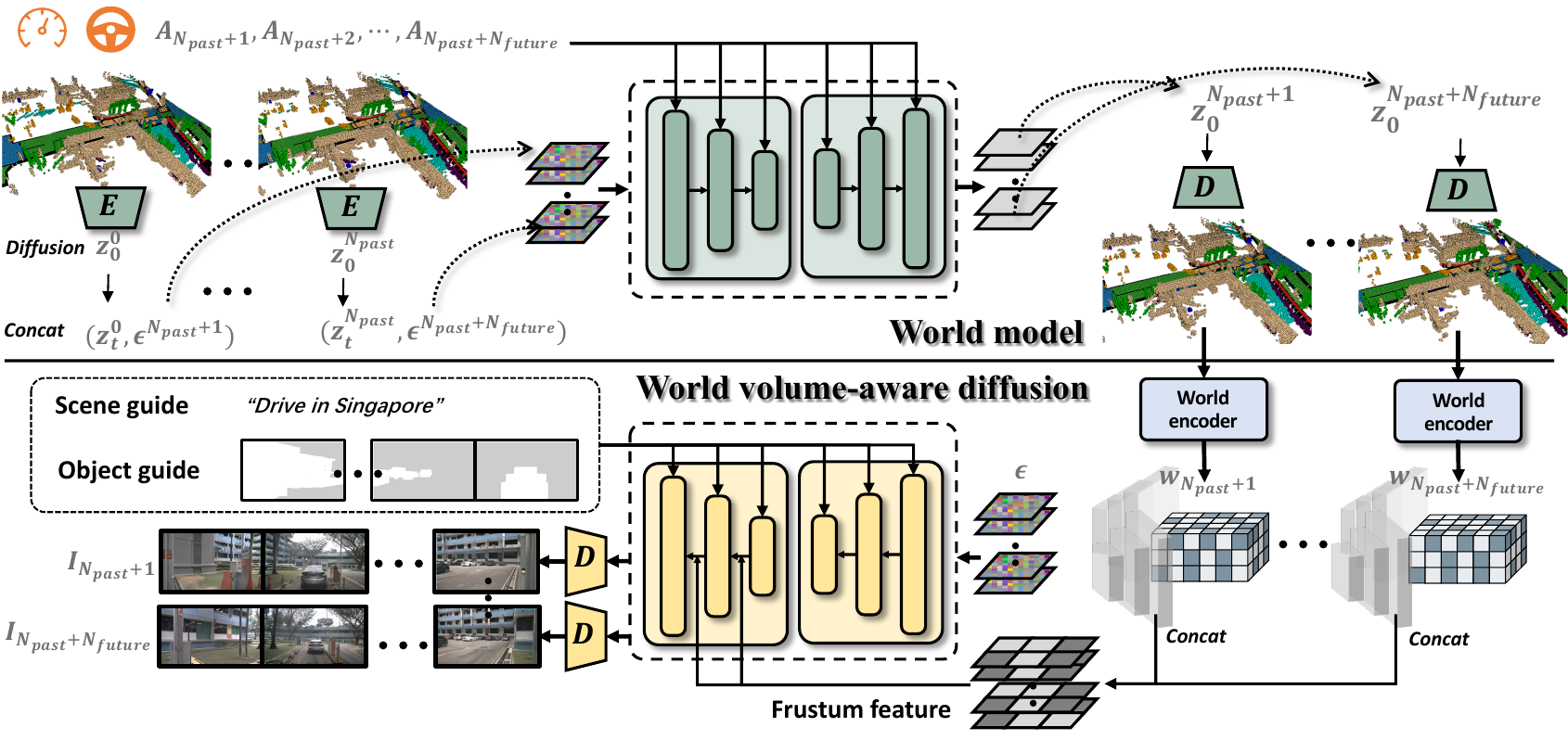} 
\caption{Overall framework of \abb. 
{\bf\em Top: world model branch.} 
We finetune the AutoencoderKL and train the 4D diffusion model from scratch to generate future world volumes based on past world volumes and the actions of the ego car.
{\bf\em Bottom: world volume-aware synthesis branch.}
Leveraging the generated future volumes as input, $\mathcal{F}_w$ are derived through the world encoder. Subsequent sampling yields $\mathcal{F}_{img}$, which are then aggregated. The process is finalized by applying panoptic diffusion to produce future videos.}
\label{fig:architecture}  
\end{figure*} 

\noindent{\bf Diffusion model for visual content generation}
{Diffusion models, as demonstrated in~\cite{ho2020denoising, song2020denoising, dhariwal2021diffusion, rombach2022high}, exhibit the capability to generate a diverse array of images through a learned denoising process.
Building upon the foundation of a well-trained image generation model, several works have emerged to augment its controllable generation capabilities, such as text-guided approaches~\cite{ding2021cogview, ramesh2022hierarchical, saharia2022photorealistic, ruiz2023dreambooth}, as well as layout-guided approaches~\cite{zhang2023adding, huang2023composer, zheng2023layoutdiffusion, li2023gligen}, etc.
Further advancements in image generation extend to video generation, emphasizing temporal consistency in recent works~\cite{singer2022make, zhou2022magicvideo, wu2023tune, blattmann2023stable}. 
Additionally, there has been a surge in interest in novel view generation techniques~\cite{liu2023syncdreamer, shi2023mvdream, liu2023zero, qian2023magic123}, focused on generating diverse perspectives from existing images. 
These advancements necessitate models with enhanced capabilities to ensure spatial consistency in generated outputs.}

{We consolidate various functionalities from preceding diffusion models and introduce \abb{}, designed specifically for generating driving scene content. 
\abb{} encompasses several innovative features: 
\textbf{(i)} Text-guided and world volume-aware controllable image generation, 
\textbf{(ii)} Spatially consistent multi-view image generation and novel view image generation, and 
\textbf{(iii)} Temporally consistent video generation.}

\noindent{\bf Rendering-based driving scene synthesis}
The rendering-based approaches~\cite{xie2023s, wu2023mars, yang2023unisim, guo2023streetsurf} initially reconstruct comprehensive world structures, followed by image or video rendering based on this reconstructed data. 
While these methods yield high-quality synthetic results, they often lack robust control capabilities.
Efforts like NeuralField-LDM~\cite{kim2023neuralfield} have attempted to introduce diffusion models into volume rendering to address diversity issues and improve control in rendering-based synthesis. 
However, they still rely heavily on numerous real-world samples from various perspectives or additional supervision during model training, raising concerns about their efficacy.

\noindent{\bf Diffusion-based driving scene synthesis}
The diffusion-based approaches~\cite{swerdlow2023street, yang2023bevcontrol, gao2023magicdrive, wang2023drivedreamer, li2023drivingdiffusion} leverage pre-trained diffusion models and incorporate sparse conditions, i.e., HD maps and bounding boxes, enabling precise control over driving scene content generation.
Part of the works~\cite{swerdlow2023street, yang2023bevcontrol, gao2023magicdrive} primarily focus on single-frame image generation. 
As a unified pipeline, they encode ground truth condition information and inject conditional features into the diffusion model to generate images corresponding to the specified condition.
Other works~\cite{wang2023drivedreamer, li2023drivingdiffusion} have ventured into driving scene video generation. 
Building upon trained single-frame generation models, they finetune the temporal module to ensure the temporal coherence of generated elements across consecutive frames in the video.

\begin{wrapfigure}{r}{0.5\linewidth}
\centering  
\includegraphics[width=1.0\linewidth]{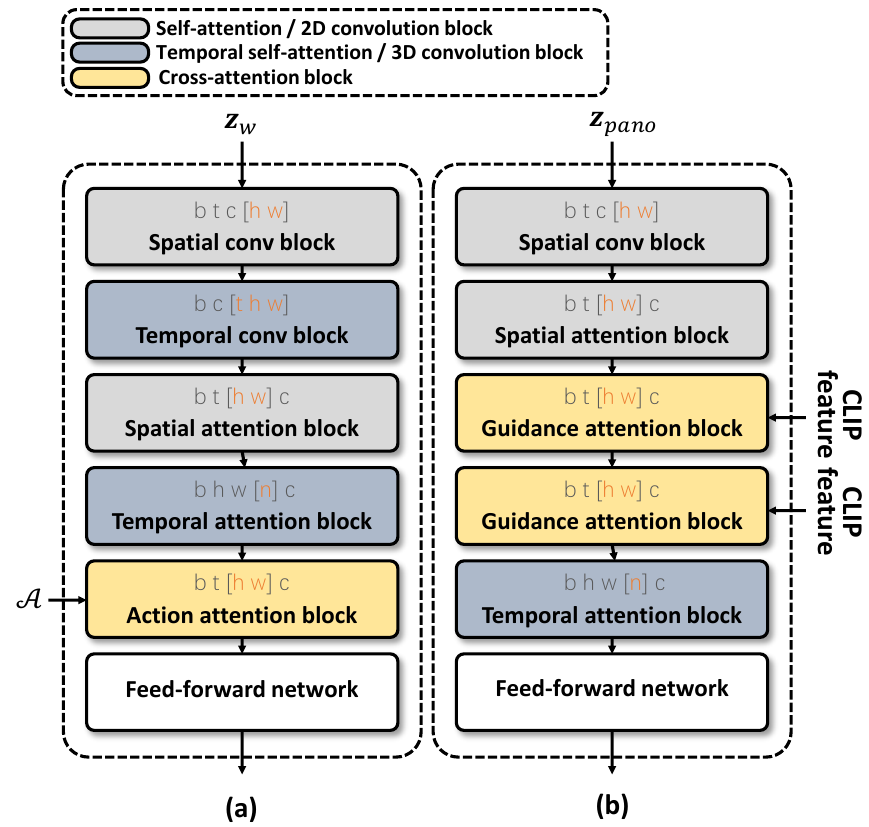} 
\caption{
{\bf (a)}: an action attention block enhances the model by incorporating action information.
{\bf (b)}: a guidance attention block integrates the CLIP feature of a specific object into the latent representation.
}
\label{fig:sub_architecture}  
\end{wrapfigure} 

The proposed \abb{} excels in generating high-quality driving scene images and videos. 
Distinguishing itself from other driving scene video generation works~\cite{wang2023drivedreamer, li2023drivingdiffusion}, \abb{} possesses unique traits: 
\textbf{(i)} It employs a self-organized dense world volume as a condition, facilitating the natural generation of geometrically consistent images from multiple perspectives. 
\textbf{(ii)} \abb{} integrates a world model capable of generating future world volumes, eliminating the necessity for ground truth conditions for each video frame generation. 
\textbf{(iii)} With the incorporation of explicit 4D information, \textit{i.e.}, temporal world volumes, \abb{} offers a comprehensive and accurate representation of the driving environment, resulting in a strong correlation with critical scene details in the generated content.
\section{Method}
\label{sec:method}
\subsection{Preliminary}
\label{sec:preliminary}
The latent diffusion model (LDM)~\cite{rombach2022high} is a generative model adept at generating high-resolution images in two stages. 
Initially, an autoencoder compresses data into a latent space (perceptual compression), where $z=E(x)$ and $\hat{x}=D(z)$ represent encoding and decoding processes, respectively. Subsequently, a denoising diffusion probabilistic model (DDPM)~\cite{ho2020denoising} models image distributions in this latent space.

The generative process reverses the diffusion sequence $\mathbf{z}_1, \ldots, \mathbf{z}_T$ using a learned Gaussian transition, formalized as:
\begin{equation}
    q\left( \mathbf{z}_{\tau} | \mathbf{z}_{\tau-1} \right) = \mathcal{N} \left( \mathbf{z}_{\tau}; \sqrt{1-\beta_\tau} \mathbf{z}_{\tau-1}, \beta_\tau \mathbf{I} \right),
\end{equation}
\begin{equation}
    p_{\theta}\left( \mathbf{z}_{\tau-1} | \mathbf{z}_{\tau} \right) = \mathcal{N} \left( \mathbf{z}_{\tau-1}; \mu_{\theta}(\mathbf{z}_{\tau}, \tau), \tilde\beta_\tau \right).
\end{equation}
Here, $\mu_{\theta}(\mathbf{z}_\tau, \tau)$ is given by a trainable noise predictor $\epsilon_{\theta}(\mathbf{z}_\tau, \tau)$.

DDPMs, viewed as a series of weight-sharing denoising autoencoders, train to predict the initial noise from a noisy input. The training goal is to maximize the variational lower bound of the negative log-likelihood:
\begin{equation}
     \mathbb{E}_{z, \epsilon, \tau} \left[ \| \epsilon - \epsilon_\theta(\mathbf{z}_{\tau}, \tau) \|^2_2 \right]. 
\end{equation}

Finally, sampling from the latent distribution and then using the latent decoder generates novel images from Gaussian noise.

\subsection{Overall architecture}
The proposed architecture, referred to as \abb{}, comprises two distinct operational branches: the {\bf\em world model branch} and the {\bf\em world volume-aware generation branch}.
The world model branch is responsible for generating future world volumes, incorporating action inputs and several initial frames of the world volume to inform its predictions. 
Meanwhile, the world volume-aware generation branch focuses on the generation of multi-camera video outputs based on temporal world volumes.
An overview of our comprehensive pipeline is visually represented in \Cref{fig:architecture}.xi

\subsection{World volume}
To effectively harness the potential of rapidly evolving generative architectures, it is imperative to transform the environmental context into a standardized format. 
Our primary focus herein lies on the extraction of high-level, abstract structural data pertinent to autonomous driving scenarios, encompassing road layouts, semantic occupancies, and other related elements.

Recognizing the critical role of three-dimensional data in rendering processes and the benefits derived from fine-grained constraints in view generation, we propose the encoding of scene information within a three-dimensional world volume, denoted as $\mathcal{W} \in \mathbb{R}^{Z \times H \times W \times C}$. 
Specifically, at any given time instance $t$, we amalgamate various facets of driving-related information to encapsulate the environmental context around the ego vehicle within a predefined range. 
This integration is achieved through the concatenation process, \textit{i.e.}, $\mathcal{W} = \text{concat}(\mathcal{O}, \mathcal{M})$, where $\mathcal{O} \in \mathbb{R}^{Z \times H \times W \times  C_{\text{occ}}}$ represents the three-dimensional semantic occupancy grid, with $C_{\text{occ}}$ indicating the number of semantic classes. 
Concurrently, $\mathcal{M} \in \mathbb{R}^{1 \times H \times W \times C_{\text{map}}}$ captures the road map information, constrained to the voxel plane at zero height.

To maintain uniformity across the height dimension, we apply zero-padding to $\mathcal{M}$ along the $Z$ axis. For a more streamlined representation, both road elements and semantic classes are encoded into the RGB spectrum, \textit{i.e.}, setting $C_{\text{map}} = 3$.

\subsection{World model}
\label{sec:world_model}
Given a clip of multi-camera videos captured during driving, denoted as 

\noindent $\{\mathcal{I}_{0}, \cdots, \mathcal{I}_{N_{\text{past}}}\}$, we initially employ off-the-shelf scene understanding models BEVDet-occ~\cite{huang2021bevdet} to infer the environment. 
This allows us to obtain the 3D world volume at these moments, represented as $\left\{\mathcal{W}_{t}\right\}_{t=1, \cdots, N_{\text{past}}}$. 
Subsequently, the world volume for the time instance $N_{\text{past}}+1$ is inferred, utilizing the past world volumes and the corresponding driving actions.

\noindent{\bf Latent world volume autoencoder}
Given the world volume $\mathcal{W} \in \mathbb{R}^{Z \times H \times W \times C}$, the encoder is designed to compress $\mathcal{W}$ into its latent representation, $z_w = E_{\mathcal{W}}(\mathcal{W}) \in \mathbb{R}^{\frac{Z}{s} \times \frac{H}{s} \times \frac{W}{s} \times C_{z}}$. 
Here, $s$ represents the downsampling factor, and is set equal to $Z$ to achieve a 2D latent representation. 
The decoder, $D_{\mathcal{W}}$, is then employed to reconstruct the original world volume from the latent representation $z_w$. 
The autoencoder is trained using a reconstruction loss and vq-regularization~\cite{esser2021taming}, aimed at minimizing the variance within the latent space.

\noindent{\bf 4D Latent world diffusion} 
After the training of the latent autoencoder, we commence the training of the world model, with both the encoder and the decoder remaining frozen. 
As illustrated in \Cref{fig:architecture}, our world model is a diffusion model that captures the conditional distribution of the latent world volume in the future, conditioned on the past world volumes and the driving actions of the ego vehicle. 
The noise predictor $\epsilon_{\phi}$ in this stage is implemented as a time-conditioned UNet~\cite{ho2020denoising}. 
The world volumes of preceding frames are first encoded by the same latent encoder to dimensions commensurate with the noised latent $z_i$, and then channel-wise concatenate with $z_i$ to serve as the input for the noise predictor.

For the driving actions, we initially tokenize the velocity $v_i$ and the steering angle $a_i$ of the ego vehicle through Fourier embedding~\cite{vaswani2017attention}, resulting in a sequence of action tokens $\mathcal{A}=\left\{v_1, a_1, \cdots, v_{N_{\text{past}}}, a_{N_{\text{past}}}\right\}$. 
These action tokens are then refined by a shallow Transformer. 
Finally, the updated action tokens are injected into the noise predictor via cross-attention layers.

\noindent{\bf Jointly modeling of the consecutive frames} 
To enhance the temporal consistency of predicted future world volumes and achieve more realistic scene dynamics, we transition to generating consecutive world volumes by considering a joint distribution of a world volume sequence.


As shown in \Cref{fig:sub_architecture}, we build upon the basic spatial Transformer block of Stable Diffusion~\cite{rombach2022high}, which comprises spatial convolution, spatial self-attention, and spatial cross-attention. Our temporal variant augments this framework by integrating temporal attention with residual connections into the spatial Transformers.

Specifically, let us denote $z_w \in \mathbb{R}^{B \times N_{\text{future}} \times C \times H \times W}$ as the sequence of latent world volumes, adding the batch dimension for clarity. The computation within the modified spatial-temporal Transformer is formalized as follows:
\begin{align*}
    \mathbf{z}_w &= \text{rearrange} \left(\mathbf{z}_w, (b\ n)\ h\ w\ c \rightarrow (b\ n)\ (h\ w)\ c  \right), \\
    \mathbf{z}_w &= \text{MHSA} \left( \text{Norm} \left( \mathbf{z}_w \right) \right) + \mathbf{z}_w, \quad \text{(Spatial)}\\
    \mathbf{z}_w &= \text{rearrange} \left(\mathbf{z}_w, (b\ n)\ (h\ w)\ c  \rightarrow (b\ h\ w)\ n\ c  \right), \\
    \mathbf{z}_w &= \text{MHSA} \left( \text{Norm} \left( \mathbf{z}_w \right) \right) + \mathbf{z}_w, \quad \text{(Temporal)}\\
    \mathbf{z}_w &= \text{rearrange} \left(\mathbf{z}_w, (b\ h\ w)\ n\ c  \rightarrow (b\ n)\ (h\ w)\ c \right), \\
    \mathbf{z}_w &= \text{MHCA} \left( \text{Norm} \left( \mathbf{z}_w, \mathcal{A} \right) \right) + \mathbf{z}_w, \quad \text{(Action)}\\
    \mathbf{z}_w &= \text{FFN} \left( \text{Norm} \left( \mathbf{z}_w \right) \right) + \mathbf{z}_w,
\end{align*}

where $\texttt{Norm}$ represents the group normalization, $\texttt{MHSA}$ stands for multi-head self-attention, $\texttt{MHCA}$ denotes multi-head cross-attention, and $\texttt{FFN}$ refers to the feed-forward network. Both $\texttt{MHSA}$ and $\texttt{MHCA}$ perform attention operations on the second dimension $l$ of the $(b \ l \ c)$ configuration.

\subsection{World volume-aware 2D feature}
With the temporally continuous world volumes $\mathcal{W}=\left \{ \W_{t} \right \}_{t=1,2,\cdots} $ generated by our diffusion-based world model, we further decode them into relational camera images for autonomous driving video generating.

\noindent{\bf World volume encoding}
The world volumes inherently possess semantic information, initially represented in the form of simplistic labels. To enhance their informativeness, we employ a featurization process utilizing CLIP~\cite{radford2021learning}:
\begin{equation}
    \mathcal{F}_w = \text{SPConv}\left(\text{PCA}\left(\text{CLIP}\left(\mathcal{W}\right)\right)\right),
\end{equation}
where $\texttt{CLIP}$ encodes the class name into the class feature. $\texttt{PCA}$ is used to decrease the dimension of the CLIP feature, thereby reducing the computational cost. $\texttt{SPConv}$ (Sparse Convolution)~\cite{spconv2022} then processes the reduced-dimensional world volume feature.
\begin{figure*}[t]  
\centering  
\includegraphics[width=\linewidth]{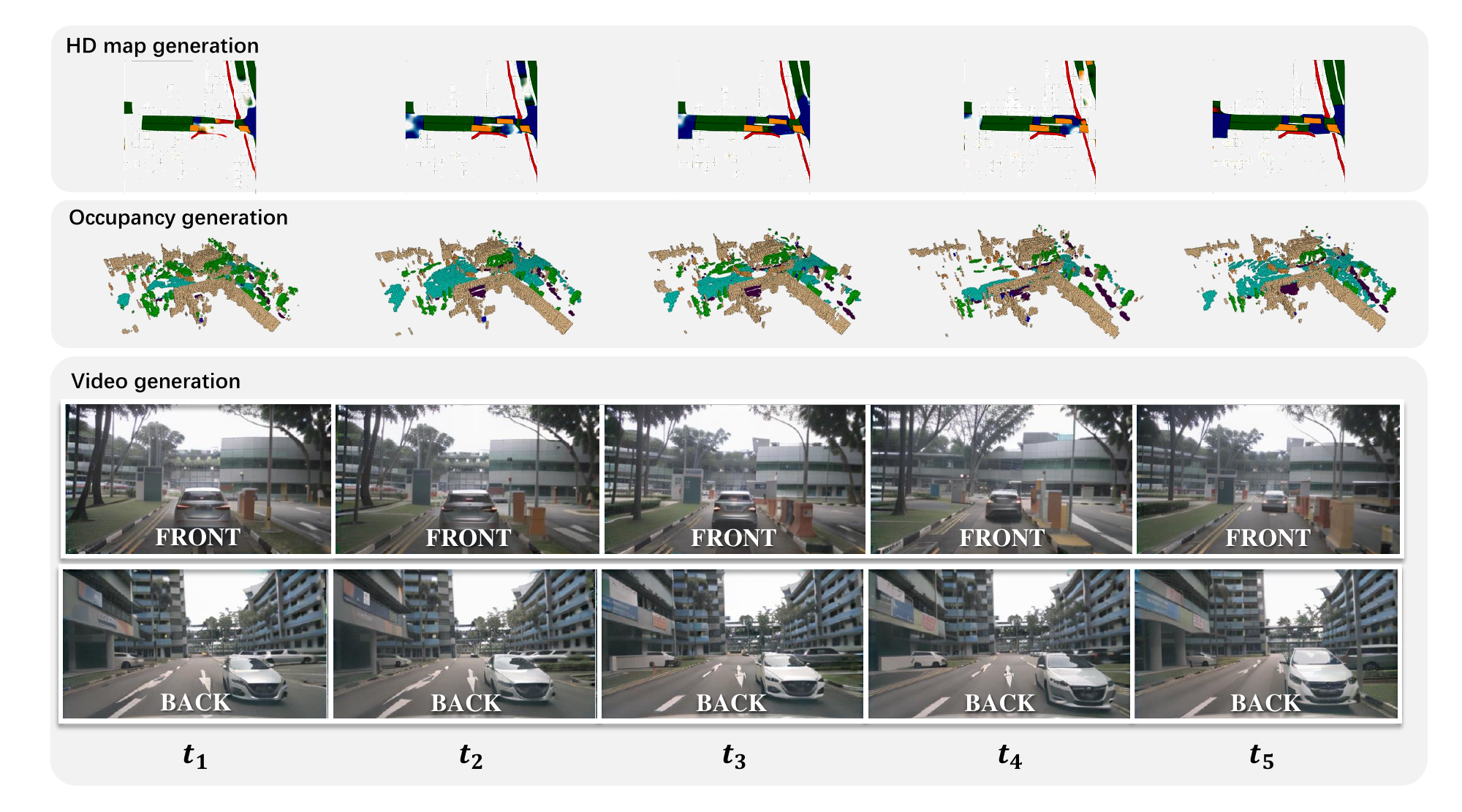} 
\caption{\abb{} excels in producing future world volumes (top two rows) with temporal consistency.
Subsequently, it utilizes the world volume-aware 2D image features derived from the world model's outputs to synthesize a driving video (bottom two rows) with both multi-camera consistency and temporal consistency.}
\label{fig:temporal}  
\end{figure*} 

\noindent{\bf Camera volume sampling}
To incorporate the world volume into image generation, we sample from it using dense rays emitted from the camera. We construct a 3D grid for each camera's frustum, denoted as $p_c$, with dimensions $D_c \times H_c \times W_c$. This grid is based on the camera's intrinsic and extrinsic properties. The process is formalized as follows:
\begin{equation}
    \mathcal{F}_{cam} = \text{interpolate}\left(p_c, \mathcal{F}_w\right),
\end{equation}
where $\mathcal{F}_{cam} \in \mathbb{R}^{B \times N_{\text{future}} \times C \times D_c \times H_c \times W_c}$ represents the camera frustum. 
We then apply a squeeze-and-excitation operation~\cite{hu2018squeeze} on the depth channel and sum along the depth dimension to obtain the world volume-aware 2D image feature:
\begin{equation}
    \mathcal{F}_{img} = \sum_{i=1}^{D_c} \text{SE}(\mathcal{F}_{cam})_{[:,\ :,\ :,\ i,\ :,\ :]}.
\end{equation}

\subsection{World volume-aware diffusion generation}
\label{sec:wvgen}
Based on the ControlNet framework~\cite{zhang2023adding}, we implement a controller that injects the aforementioned world volume-aware 2D image feature $\mathcal{F}_{\text{img}}$ into the pretrained latent diffusion model.
The Transformer architecture of the UNet is illustrated in \Cref{fig:sub_architecture}. However, we initially omit the Temporal Attention Block when training the single-frame Generation model.

\noindent{\bf Panoptic diffusion}
We aggregate the world volume-aware 2D image feature from different view into a single panoptic feature $\mathcal{F}_{pano}$ input to the diffusion model:
\begin{equation}
\mathcal{F}_{pano} = \begin{bmatrix}
\mathcal{F}_{img}^{\text{front left}} & \mathcal{F}_{img}^{\text{front}} & \mathcal{F}_{img}^{\text{front right}}\\
 \mathcal{F}_{img}^{\text{back right}} & \mathcal{F}_{img}^{\text{back}} & \mathcal{F}_{img}^{\text{back left}}
\end{bmatrix}.
\end{equation}
Naturally, the decoding target is transformed into the corresponding panoptic image.
This operation shifts the focus from ensuring consistency between latent codes for different views to maintaining inherent consistency within a single latent code $z_{pano}\in \mathbb{R}^{C\times 2H_c\times 3W_c}$, resulting in a unified and coherent appearance in multi-camera image generation.

\noindent{\bf Scene guidance}
Except for the conditional feature, we also introduce text prompt-based scene guidance into the latent representations. 
This involves extracting the CLIP feature from a text-based description of an image and mapping this text feature to the intermediate layers of the latent diffusion model, following a similar approach as LDM~\cite{rombach2022high}.
 
\noindent{\bf Object guidance}
When it comes to objects that require specific placement within the generated image, we emphasize their pixel-level locations in the latent space using a cross-attention calculation.

Specifically, we employ the voxel-based projection of the initial world volume onto each camera plane, resulting in the creation of preliminary masks $m_{\text{class}}$ distinguished by the world volumes' semantic categories. Then, cross-attention is calculated by:
\begin{equation*}
    \mathbf{z}_{pano} =  \text{MHCA}(\mathbf{z}_{pano}(m_{\text{class}}=1), \text{CLIP}(\text{class})) + \mathbf{z}_{pano},
\end{equation*}
where $\text{CLIP}(\text{class})$ represents the category-specific CLIP feature, with semantic categories \texttt{class} being defined as bus, car, pedestrian, truck, construction, and vegetation.

\subsection{Video generation}
Before we generate video, we first train the single-frame multi-camera generation model, and then we add temporal attention block and only train the temporal attention block as shown in the right side of \Cref{fig:sub_architecture} and freeze the other blocks.

\begin{figure*}[h]  
\centering  
\includegraphics[width=\linewidth]{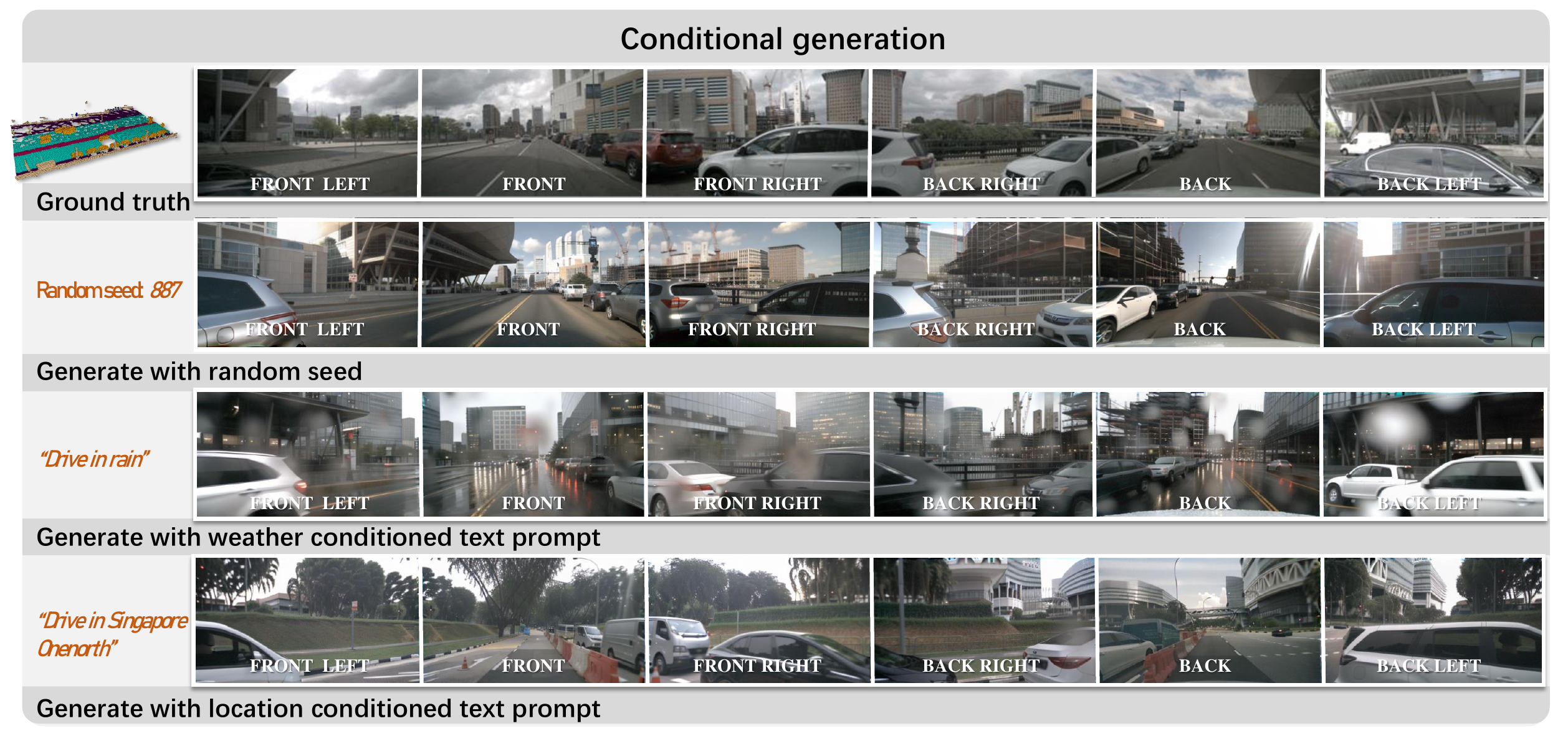} 
\caption{Examples of conditional generation on nuScenes~\cite{caesar2020nuscenes} validation dataset. \abb{} empowers diverse and controllable scene generation.
Altering the random seed allows for the generation of various scenarios.
Additionally, adjustment to weather (such as sunny, rainy, night, etc.) and location (Singapore, Boston, etc.) within the prompt enables the modification of weather conditions and city styles within the generated scene.
}
\label{fig:control_gen}  
\end{figure*} 

\noindent{\bf Temporal consistency}
The temporal attention block is added to ensure the generation of multi-frame images consistent, which calculates the attention by:
\begin{align*}
    \mathbf{z}_{pano} &= \text{rearrange} \left(\mathbf{z}_{pano}, (b\ n)\ (h\ w)\ c  \rightarrow (b\ h\ w)\ n\ c  \right), \\
    \mathbf{z}_{pano} &= \text{MHSA} \left( \text{Norm} \left( \mathbf{z}_{pano} \right) \right) + \mathbf{z}_{pano}, \quad \text{(Temporal)}
\end{align*}


\begin{figure}[!htb]  
\centering  
\includegraphics[width=\linewidth]{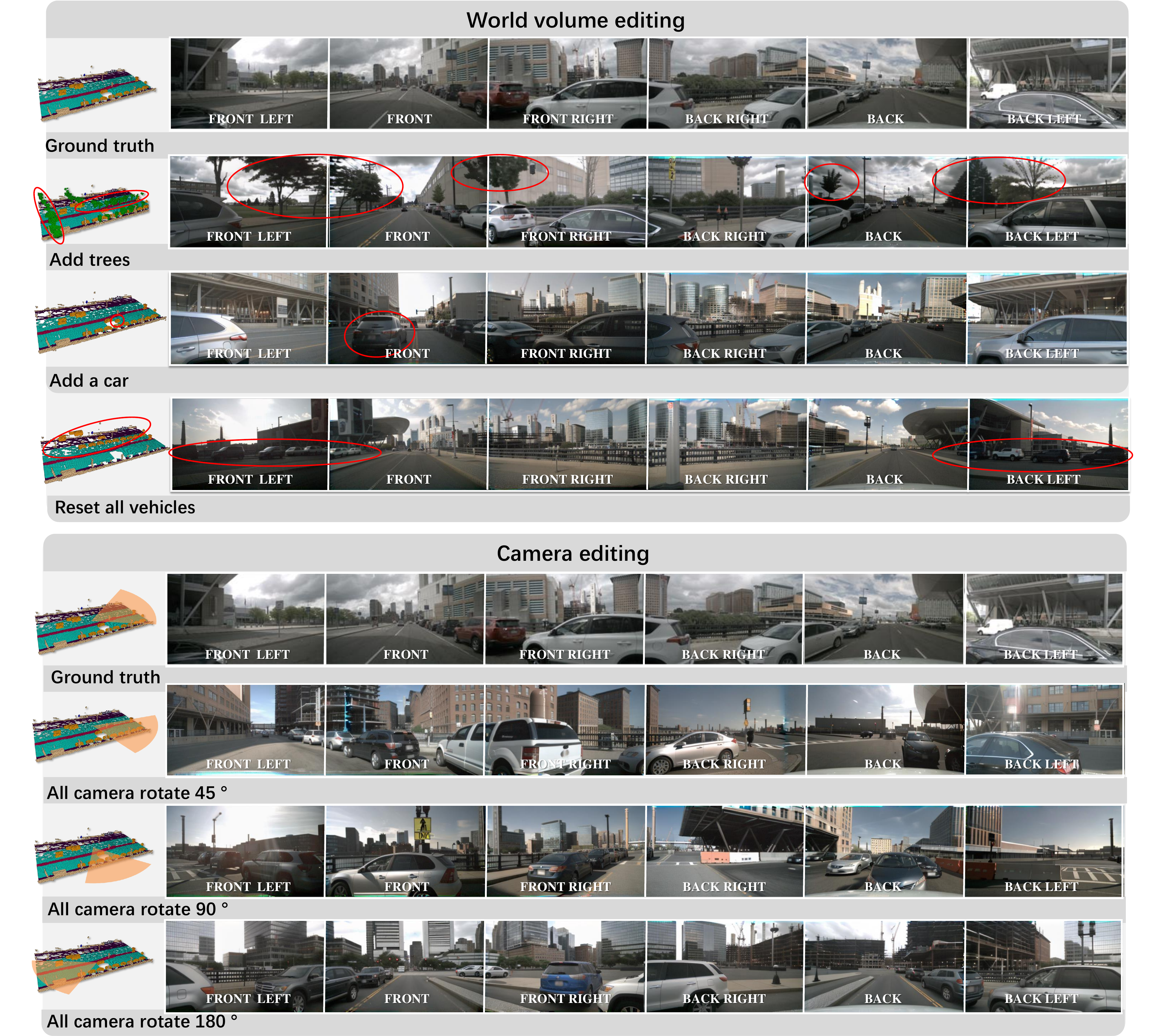} 
\caption{Examples of controlled editing on nuScenes~\cite{caesar2020nuscenes} validation dataset.
{\bf \em Top:} the ability to selectively add or remove specific objects (such as trees, buildings, cars, etc., highlighted by red circles in the figure) within the world volume empowers the precise and coherent generation of diverse driving scenarios across multiple cameras.
{\bf \em Bottom:} due to \abb{}'s advanced 3D understanding capabilities, the rotation of perspectives across multiple cameras can be achieved by modifying the camera's extrinsic parameters. 
}
\label{fig:control_edit}  
\end{figure} 
\section{Experiments}
\label{sec:exp}

\subsection{Experimental setup}
\noindent{\bf Dataset}
We conducted experiments using the nuScenes~\cite{caesar2020nuscenes} dataset which comprises 700 training videos and 150 validation videos. 
Our study utilized images from the six camera views.

For ground truth occupancy, we use the data from CVPR 2023 3D Occupancy Prediction Challenge~\cite{tong2023scene, tian2023occ3d}, aligned with semantic labels consistent with the nuScenes-lidarseg dataset~\cite{fong2022panoptic}. 
The ego car-centric local map is derived from the global HD map using the vehicle's ego pose and precisely aligned with the occupancy data. 

\noindent{\bf Data preprocessing}
Images are resized to a resolution of 256 $\times$ 448, and six consecutive frames are selected for video training.
Text prompts are constructed using a structured template: ``\textit{Drive in \{weather description\} in \{location\}. 
The driving scene is in \{environment description\}, captured by multi-view camera.}''
The descriptive details enclosed in brackets are filled using scene descriptions from nuScenes.

\noindent{\bf Training}
The training process utilized eight NVIDIA A6000 GPUs, encompassing 30,000 iterations for the world volume auto-encoder, 50,000 iterations for volume diffusion, 50,000 iterations for single-frame generation training, and 3,500 iterations for refining video generation.

\noindent{\bf Inferring}
We begin with three initial world volumes organized from ground-truth data and proceed to generate the subsequent three world volumes by incorporating actions as input. 
This iterative process continues, producing a quasi-long world volume video sequence. 
Ultimately, the world volume video sequence is decoded to form a multi-view camera video.

\noindent{\bf Evaluation}
We conducted a comparative analysis of our method's image quality against other approximate works in the following text. 
To quantitatively assess our work, we employed established metrics such as Fréchet Inception Distance~\cite{heusel2017gans} (FID) and Fréchet Video Distance~\cite{unterthiner2018towards} (FVD) for evaluating the quality of the generated video.
Additionally, we provided qualitative demonstrations that underscore the advantages of our proposed method.

\subsection{Results}
\subsubsection{4D World volume generation}
The distinctive feature of \abb{} lies in its ability to predict high-level driving environments in the first stage, including occupancy and HD maps, as vividly illustrated in the left two columns of \Cref{fig:video_teaser} and the top two rows of \Cref{fig:temporal}. 
Given vehicle actions from the dataset, \abb{} successfully foresees high-quality future driving environments. 
Notably, our model retains its effectiveness in predicting single frames while ensuring a high degree of consistency across the generated future frames. 
This dual capability of \abb{} — excelling in both short-term prediction and long-term scene evolution — highlights its potential as a powerful tool for advanced driving environment simulation.

\subsubsection{Multi-camera single-frame image generation}
\Cref{fig:control_gen} shows the qualitative results of the multi-camera single-frame image generation under diverse conditions.
We can observe that the generated images have highly consistency.
Furthermore, from the first row of \Cref{fig:control_gen}, we can see that the generated images accurately correspond to the occupancy guide. 
Beyond that, we can also control the style of the generated images through natural language.
In the last two rows of the \Cref{fig:control_gen}, our \abb{} demonstrates a profound comprehension of the specified textual conditions, skillfully creating highly realistic simulations of rainy day scenarios. 
When prompted with ``\textit{Drive in Singapore Onenorth,}'' the model skillfully generates a scene embodying the garden city essence typical of that region. 

The quantitative results in \Cref{tab:eval} underscore that \abb{} showcases markedly lower FID (27.6) compared to DriveGan \cite{kim2021drivegan} and DriveDreamer \cite{wang2023drivedreamer}, while these methods are limited to single-view camera image generation. 
This result solidifies the improved capacity of our approach in generating more realistic autonomous driving images.

\subsubsection{Multi-camera single-frame image editing}
Given the world volume-aware character of \abb{}, we can realize editing based on the edited world volume.

\noindent{\bf Rearranging objects:} We can add, delete, or transform the objects within the occupancy. 
As shown in \Cref{fig:control_edit}, we can add trees, vehicles, or rearrange the location of objects.

\noindent{\bf Camera extrinsic editing:} More interestingly, we can rotate the camera stereo when projecting the world volume to the camera volume and generate the corresponding scene under a new camera setting.

\renewcommand\arraystretch{1.2} 
\begin{table}[htb] 
    \scriptsize
	\centering  
 \caption{Quantitative comparison of image/video generation quality on Nuscenes validation set. \abb{} achieve both multi-view and multi-frame generation, demonstrating the lowest FID and FVD scores among all methods.}
	\begin{tabular}{r||c|c||c|c} 
 \hline

 \hline
    \rowcolor{gray!20}
    \textbf{Method} & \textbf{Multi-view} & \textbf{Multi-frame} & \textbf{FID$\downarrow$} & \textbf{FVD$\downarrow$}\\
	\hline
    DriveGan~\cite{kim2021drivegan}    & $\surd $ &  & 73.4 & 502.3 \\   
    DriveDreamer~\cite{wang2023drivedreamer}   &  & $\surd $ & 52.6 & 452.0 \\ 

    \hline
    Ours(single-frame)   & $\surd $ &  & 27.6 & - \\ 
    Ours(video)   & $\surd $ & $\surd $ & - & 417.7  \\ 
    
	\hline

 \hline
	\end{tabular}
         
	\label{tab:eval}
\end{table}

\subsubsection{Multi-camera video generation}
\Cref{fig:temporal} showcases the qualitative results of the multi-camera video generation.
Thanks to the high-quality world volume generation and finetuning for temporal consistency, the generated video demonstrates a remarkable level of temporal coherence. 
The appearance of objects and background within the video maintains consistency throughout the sequence.
Furthermore, the quantitative results of video quality, indicated by the FVD (417.7) in \Cref{tab:eval}, underscores the superiority of our method compared to the other works.

\subsection{Ablation studies}
\subsubsection{Effectiveness of world volume and object guidance}
In this study, we examine the significance of world volume and object guidance, as detailed in \Cref{tab:obg_ab}. We analyze three scenarios: object guidance alone, world volume alone, and the combination of world volume with object guidance. When solely relying on object guidance, interpreted as object mask information, the generated results are notably inferior. In contrast, incorporating world volume, regardless of the presence of object guidance, significantly enhances the generation outcomes. This underscores the critical importance of world volume in the generative process. Additionally, the incorporation of object guidance leads to more precise object localization, especially when evaluating the fidelity of the generated images, as indicated by the highest FID scores.

\subsubsection{Consistency}
To assess the quantitative outcomes of multi-camera and temporal consistency, we adopt a quantitative methodology utilizing Key Points Matching (KPM)~\cite{wang2023driving}. This approach involves a pre-trained matching model~\cite{sun2021loftr} to calculate the average number of matching keypoints. \Cref{tab:obg_ab} reveals that there is an impressive match rate of up to 80.2\% for keypoints across successive frames, underscoring a significant accomplishment in maintaining temporal consistency. Regarding multi-view consistency, we observe that 90.0\% of keypoints match between adjacent cameras, further highlighting our methodology's efficacy in ensuring robust multi-camera consistency.

\begin{table}[htb] 
	\centering  
 \caption{{\bf\em Left:} Quantitative comparison of image generation quality on the nuScenes validation set with and without object guidance. {\bf\em Right:} Quantitative results of multi-view consistency and tmerpoal consistency. KPM is used for evaluation}
	\begin{tabular}{cc|c cc|c} 
	\cline{0-2} \cline{5-6} 

    World Volume & Object Guidance & \textbf{FID$\downarrow$} & \ \ \ \  & Consistency & \textbf{KPM}(\%)$\uparrow$\\
	\cline{0-2} \cline{5-6}

    $\times$ & $\checkmark$ & 28.6 && Multi-view & 90.0\\ 
    $\checkmark$ & $\times$ & 27.8 && Temporal & 80.2 \\ 
    \cline{5-6} 
     $\checkmark$ & $\checkmark$ & \textbf{27.6}\\
    
	\cline{0-2} 
	\end{tabular}
         
	\label{tab:obg_ab}
\end{table}

\section{Conclusion}
\label{sec:conclusion}
In this paper, we propose \abb{}, which marks a significant advancement in generating multi-camera driving scene videos. 
Leveraging a novel 4D world volume, it adeptly integrates time with spatial data, addressing the complexity of creating content from multi-sensor data while ensuring consistency. 
This two-phase system not only produces high-quality videos based on vehicle controls but also enables complex scene editing, showcasing its potential as a comprehensive tool for advancing autonomous driving technologies.

\clearpage  

\section*{Acknowledgments}
This work was supported in part by National Natural Science Foundation of China (Grant No. 62106050 and 62376060),
Natural Science Foundation of Shanghai (Grant No. 22ZR1407500), 
USyd-Fudan BISA Flagship Research Program and Lingang Laboratory (Grant No. LG-QS-202202-07).

%
%
\bibliographystyle{splncs04}
\bibliography{main}

\clearpage
\appendix
\section{Appendix}
\subsection{Masks for object guidance}
We have devised an efficient approach to compute masks for object guidance.

\begin{figure*}[h]  
\centering  
\includegraphics[width=\linewidth]{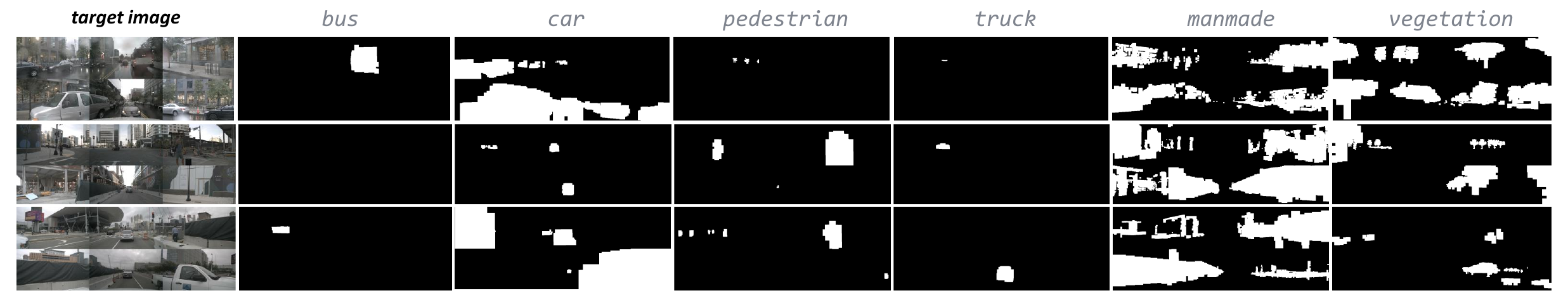} 
\caption{\textbf{Object masks calculated by voxel projection.} Origin images and per-class masks are organized as: \textit{\textbf{Top:}} front left, front, front right; \textit{\textbf{Bottom:}} back right, back, back left.}
\label{fig:mask_gen}  
\vspace{-0.2cm}
\end{figure*}

Specifically, for the original occupancy voxel grids belongs to a certain class $P_{class} \in \mathbb{R}^{N\times3}$, our initial step involves projecting them onto the camera imaging plane using the camera's intrinsic properties $K$ and extrinsic properties $T$, performed as follows:
\begin{equation}
    P_{proj}=K(RP_{class}+t),
\end{equation}
where $R=T^{-1}[:3,:3]$ and $t=T^{-1}[:3,3]$ represent the rotation matrix and translation vector of the inverse of $T$, respectively. 
Following this, we initialize an all-zero image $m_{class}$, matching the dimensions of the camera image. 
For each normalized projected point $P_{proj}^{i}$, we efficiently simulate voxel projections by marking a square-shaped region in $m_{class}$ as occupied:
\begin{equation}
    m_{class}[P_{proj}^{i_{x}}-\delta:P_{proj}^{i_{x}}+\delta,P_{proj}^{i_{y}}-\delta:P_{proj}^{i_{y}}+\delta]=1,
\end{equation}
where $P_{proj}^{i_{x}}$ and $P_{proj}^{i_{y}}$ represent the normalized x and y components of $P_{proj}^{i}$, while $\delta$ is inversely proportional to the depth of the projection points $P_{proj}^{i_{z}}$:
\begin{equation}
    \delta=d/P_{proj}^{i_{z}},
\end{equation}
where $d$ is a hyperparameter set to 375.
Several generated mask samples are illustrated in \Cref{fig:mask_gen}.

\subsection{Zero-shot generation}
\begin{figure*}[!htb]  
\centering  
\includegraphics[width=0.9\linewidth]{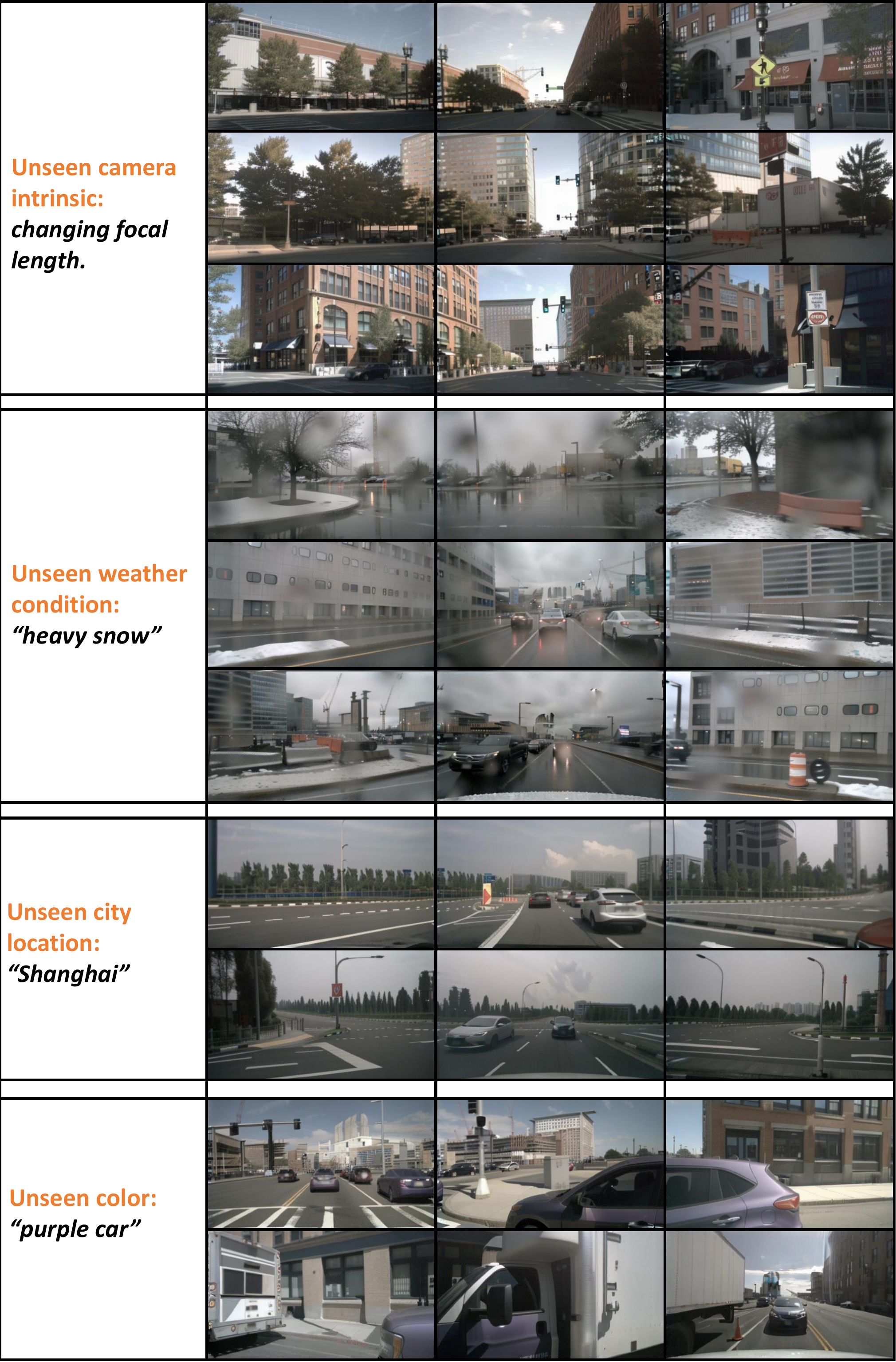} 
\caption{Examples of out-of-domain generation}
\label{fig:outofdomain}  
\end{figure*}

In Figure~\ref{fig:outofdomain}, we present intriguing findings regarding zero-shot generation outcomes.

Beyond modifying the camera's extrinsic parameters, we also experimented with changing the intrinsic settings of the camera, configurations not encountered in the dataset. The successful generations demonstrate our model's ability to generalize across different world volumes, indicating that it genuinely learns from world volume features instead of merely overfitting to the data.

We experimented with an unseen weather condition, "heavy snow," by altering the weather condition component of the scene guidance. This snow weather condition is entirely absent in the nuScenes dataset~\cite{caesar2020nuscenes}. Nonetheless, our model incorporates the concept of snow from the Stable Diffusion pretrained model, integrating it into the scene based on its understanding of snow weather. This demonstrates the model's ability to amalgamate concepts from the pretrained model with the fine-tuned street view images, showcasing its versatility and adaptability in generating content under novel conditions.

Interestingly, we altered the location word to the entirely different city of Shanghai, which presents significant differences from Singapore and Boston. This modification resulted in the generation of images that depict foggy-like weather and the architectural style characteristic of Shanghai's highways. This ability to adapt the visual representation to a distinct geographical context further illustrates the model's capacity to inherit and apply concepts from the Stable Diffusion pretraining. It showcases the model's remarkable flexibility in synthesizing elements that reflect specific environmental and cultural attributes associated with various locations.

We also explored out-of-domain modifications in object guidance by substituting the word "car" with "purple car." Purple cars are quite rare in the dataset. Nonetheless, our model effectively integrated the concept of purple, successfully altering all cars in the scene to purple. This experiment underscores the model's ability to comprehend and apply color attributes to specific objects within a scene, demonstrating its proficiency in adapting to and implementing novel characteristics not commonly found in the training data.

\subsection{Downstream tasks}
To enhance performance through more comprehensive data training, we integrate our generation results with the nuScenes~\cite{caesar2020nuscenes} training set as a supplementary data source.
We test our results on 3D object detection, with BEVDet~\cite{huang2021bevdet} as baseline.
We evaluate our outcomes in 3D object detection, using BEVDet as our baseline model. 
Our approach involves generating a complete training dataset based on the ground truth. Subsequently, we train the 3D object detection model using BEVDet on both the original nuScenes dataset and our newly generated nuScenes dataset.
Since our world volume class encompasses only cars, trucks, and buses, our testing focuses exclusively on vehicles within these categories.

\begin{table*}[htb] 
	\centering  
	\begin{tabular}{l|ccccccc} 
\hline

\hline
    \rowcolor{gray!20}
    \textbf{Method} & 
    \textbf{NDS$_\text{v}\uparrow$} &  \textbf{mAP$_\text{v}\uparrow$} &  
    \textbf{mATE$_\text{v}\downarrow$} &
    \textbf{mASE$_\text{v}\downarrow$} &
    \textbf{mAOE$_\text{v}\downarrow$} &
    \textbf{mAVE$_\text{v}\downarrow$} &
    \textbf{mAAE$_\text{v}\downarrow$} \\
	\hline

    original   & 17.45 & 34.9 &69.2 & 20.4 & 17.9 & 124.6 & 28.6\\ 
    + generated  & \textbf{18.10} & \textbf{36.2} & \textbf{68.6} &\textbf{20.1}& \textbf{15.7} & \textbf{123.4} & \textbf{28.1}\\ 
    
	\hline

\hline
	\end{tabular}
         \caption{The enhancement brought by our generated data for 3D object detection. The evaluation is on the vehicle classes of cars, trucks, and buses.}
	\label{tab:downstream}
\end{table*}

As indicated in Table~\ref{tab:downstream}, we observed a significant improvement in the mean Average Precision (mAP). 
Upon analyzing the error contributions, it's evident that the most substantial improvement arises from a reduction in orientation error, which decreased from 17.9 to 15.7. 
This demonstrates that our generated data indeed enhances the training of downstream tasks, such as 3D object detection.

\subsection{Ablation studies}

\begin{figure*}[!htb]  
\centering  
\includegraphics[width=\linewidth]{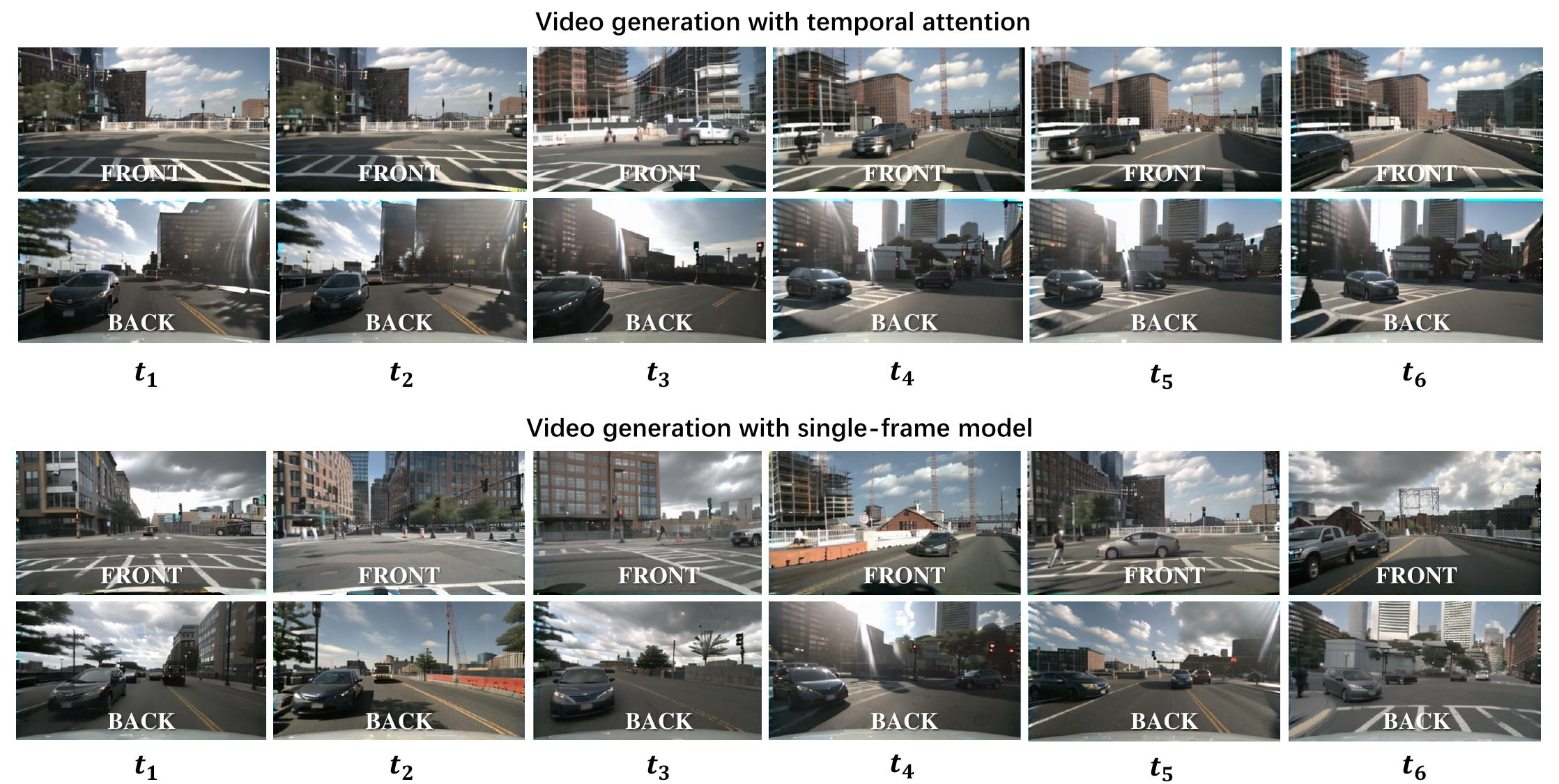} 
\caption{\textbf{Qualitative ablation study on temporal finetuning.} Consistency in object and background preservation across frames is evident when conduct temporal finetuning.} 
\label{fig:temporal_ab}  
\vspace{-0.2cm}
\end{figure*}

\paragraph{Effectiveness of temporal finetuning}
To verify the effectiveness of the video consistency module, we illustrate the simulation outcomes of the front and back cameras in \Cref{fig:temporal_ab} after removing the temporal finetuning.

Obviously, the finetuned model notably maintains consistent object and background appearances across frames. 
In contrast, the single-frame model struggles to achieve this level of consistency.

\subsection{Video results}
We show the video results in file \texttt{10410\_video.mp4}.

\subsection{Additional single-frame image generation results}
\begin{figure*}[h]  
\centering  
\includegraphics[width=\linewidth]{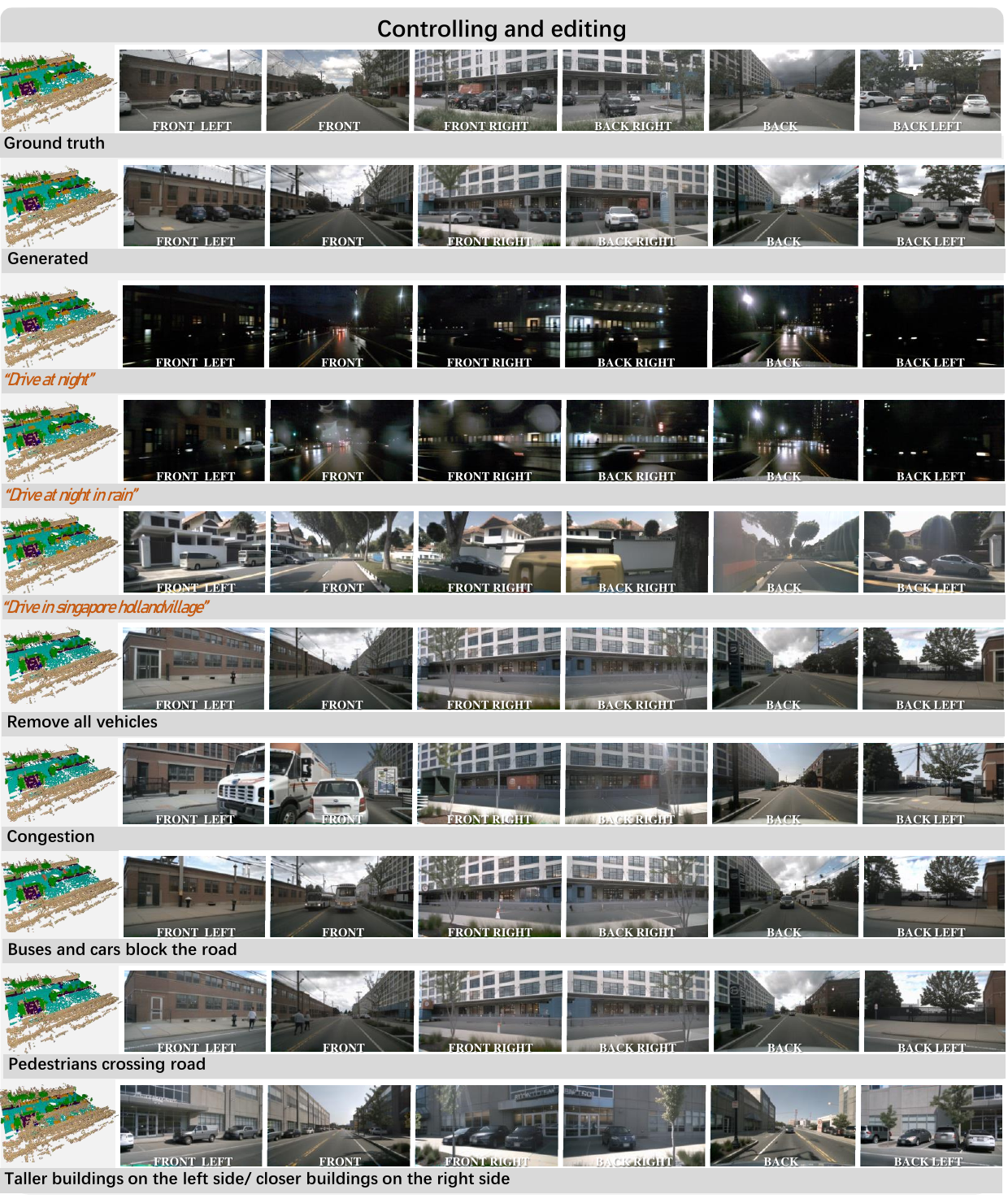} 
\caption{\textbf{Additional controlling and editing samples.}} 
\label{fig:add_gen}  
\vspace{-0.2cm}
\end{figure*}

We provide additional single-frame image generation samples showcasing more diverse editing and controlling capabilities in \Cref{fig:add_gen}.

Utilizing the Lego-style edit-friendly features of the world volume offers boundless possibilities for editing and controlling, resulting in numerous potential generation outcomes, some of which may even be rare in the real world.

\begin{figure*}[h]  
\centering  
\includegraphics[width=\linewidth]{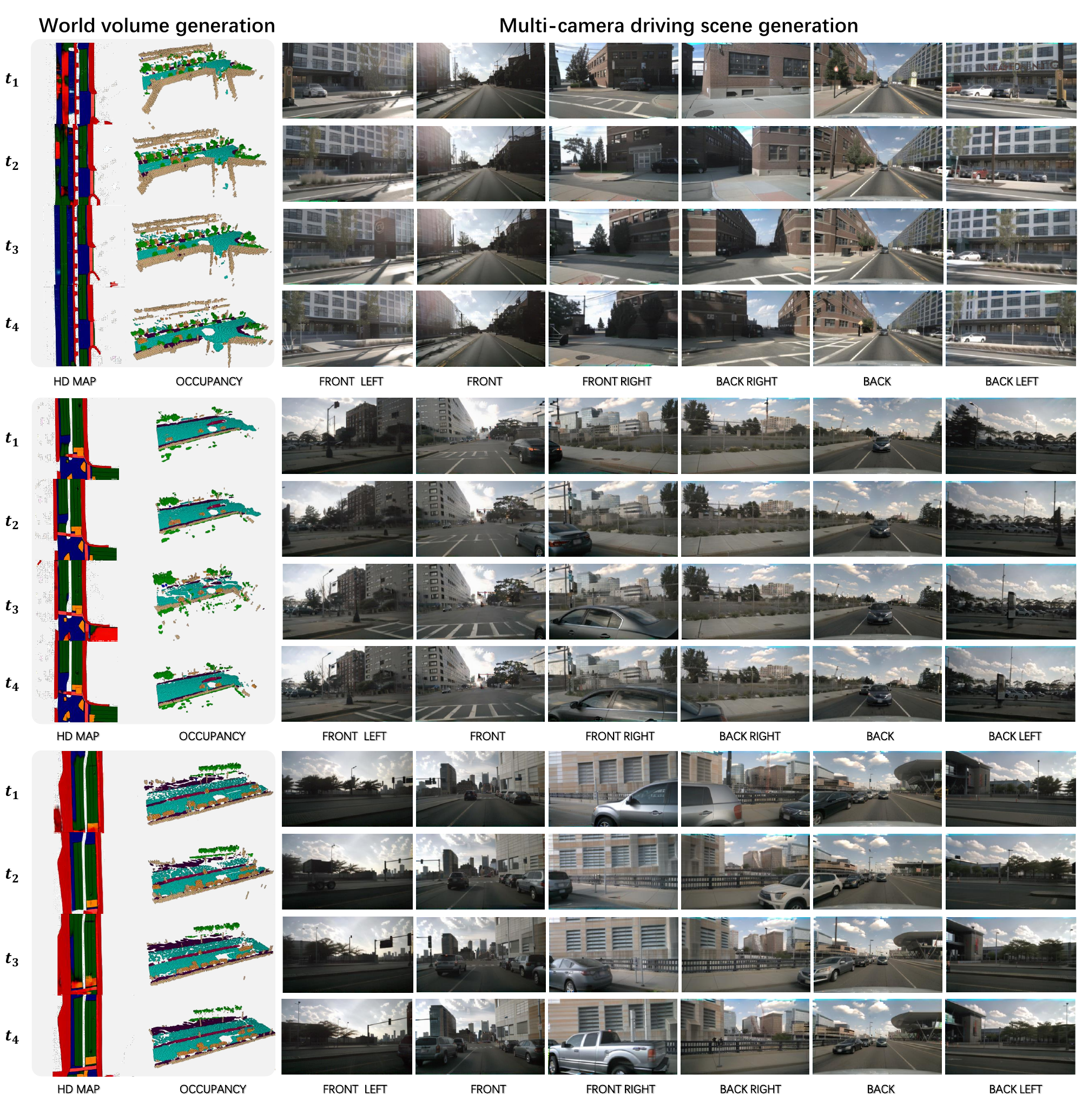} 
\caption{\textbf{Additional videos generated from the generated world volumes}} 
\label{fig:add_video_1}  
\vspace{-0.2cm}
\end{figure*}

\begin{figure*}[h]  
\centering  
\includegraphics[width=\linewidth]{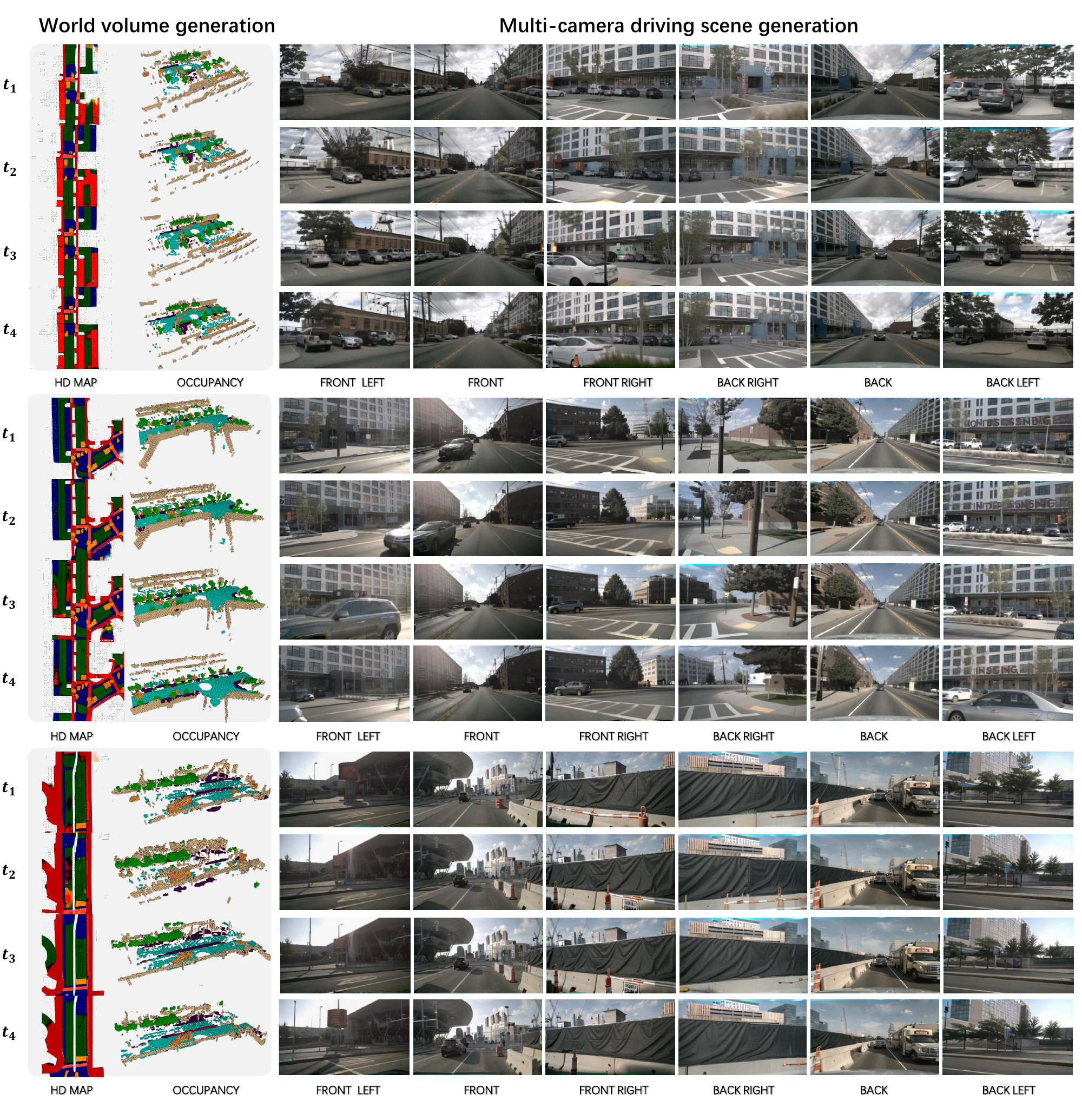} 
\caption{\textbf{Additional videos generated from the generated world volumes}} 
\label{fig:add_video_2}  
\vspace{-0.2cm}
\end{figure*}

\begin{figure*}[h]  
\centering  
\includegraphics[width=\linewidth]{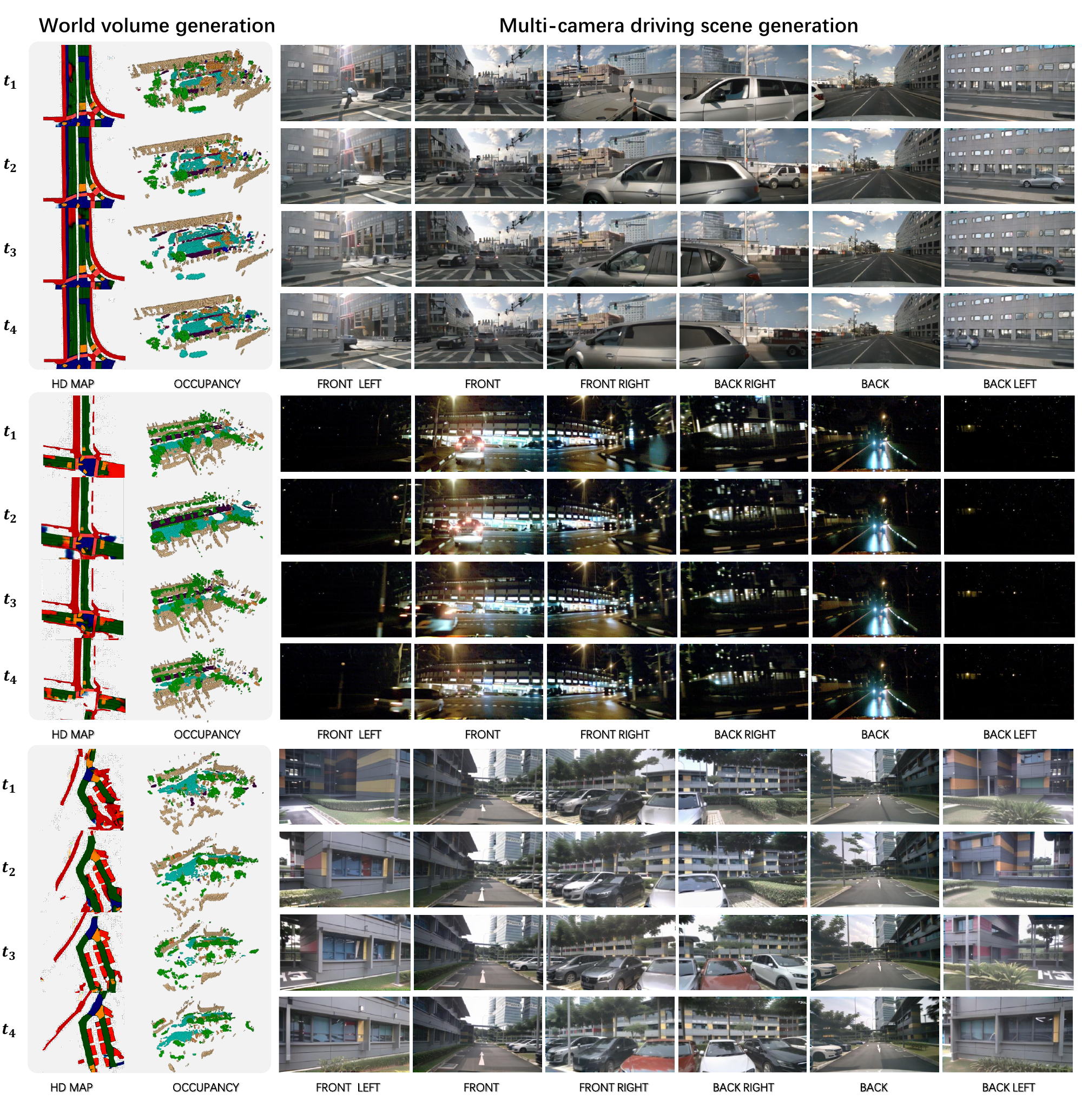} 
\caption{\textbf{Additional videos generated from the generated world volumes}} 
\label{fig:add_video_3}  
\vspace{-0.2cm}
\end{figure*}

\begin{figure*}[h]  
\centering  
\includegraphics[width=\linewidth]{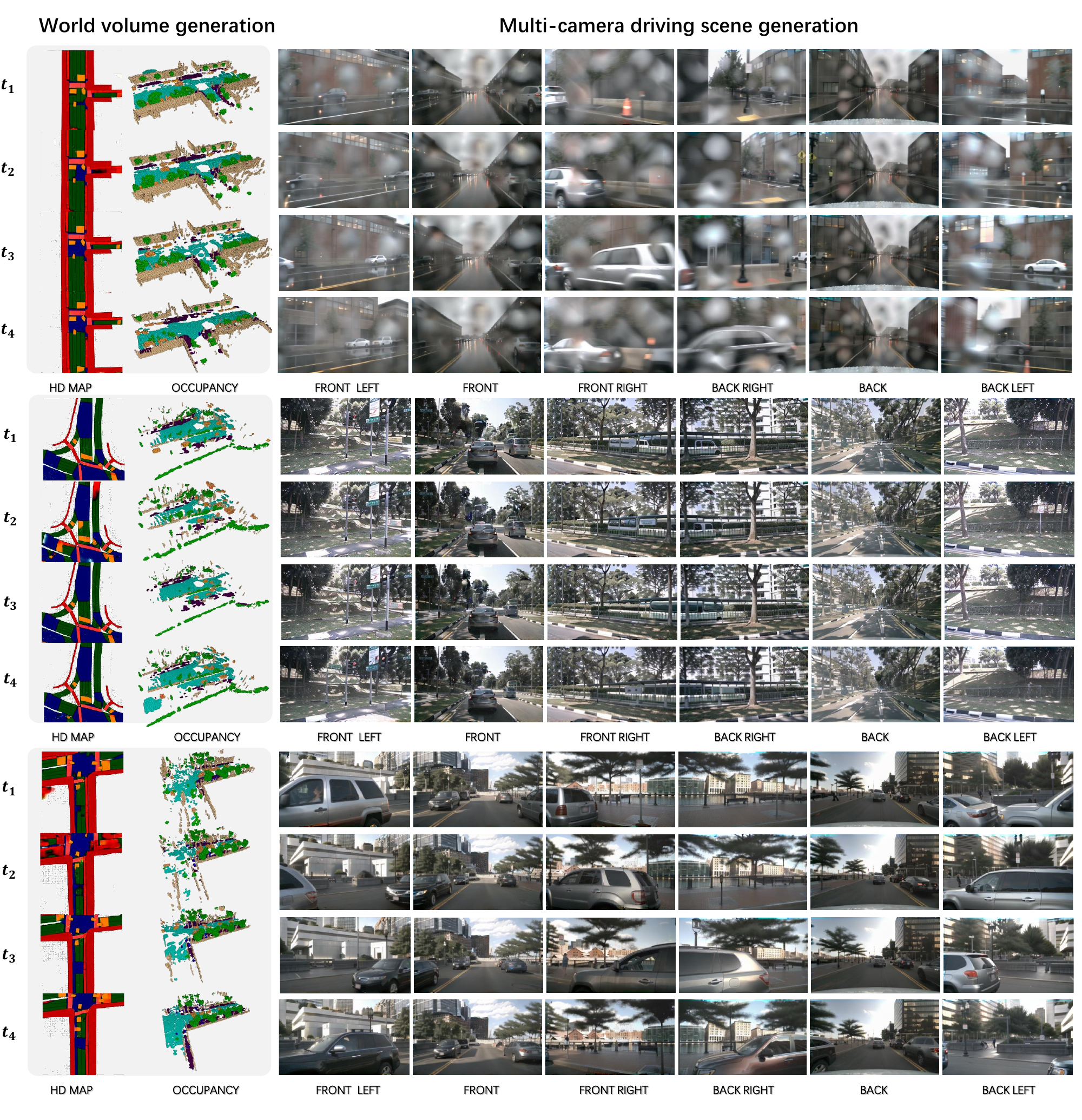} 
\caption{\textbf{Additional videos generated from the generated world volumes}} 
\label{fig:add_video_4}  
\vspace{-0.2cm}
\end{figure*}

\end{document}